\documentclass[runningheads]{llncs}
\usepackage{graphicx}
\usepackage{amsmath}
\usepackage{amssymb}
\DeclareMathOperator*{\argmin}{arg\,min}
\DeclareSymbolFont{matha}{OML}{txmi}{m}{it}
\DeclareMathSymbol{\varv}{\mathord}{matha}{118}
\usepackage{array}
\usepackage{tabularx}
\usepackage{multirow, booktabs}
\usepackage{float}
\usepackage{bm}
\usepackage[breaklinks=true,bookmarks=false]{hyperref}

\begin{document}
%
\title{Conditional Deformable Image Registration with Convolutional Neural Network}
%
\titlerunning{Conditional Deep Deformable Image Registration with CNN}
%
%
%
%
\author{Tony C. W. Mok \and Albert C. S. Chung}

\institute{
	Department of Computer Science and Engineering,\\
	The Hong Kong University of Science and Technology, Hong Kong\\
	\email{\{cwmokab,achung\}@cse.ust.hk}
}

\maketitle              
\begin{abstract}
Recent deep learning-based methods have shown promising results and runtime advantages in deformable image registration. However, analyzing the effects of hyperparameters and searching for optimal regularization parameters prove to be too prohibitive in deep learning-based methods. This is because it involves training a substantial number of separate models with distinct hyperparameter values. In this paper, we propose a conditional image registration method and a new self-supervised learning paradigm for deep deformable image registration. By learning the conditional features that are correlated with the regularization hyperparameter, we demonstrate that optimal solutions with arbitrary hyperparameters can be captured by a single deep convolutional neural network. In addition, the smoothness of the resulting deformation field can be manipulated with arbitrary strength of smoothness regularization during inference. Extensive experiments on a large-scale brain MRI dataset show that our proposed method enables the precise control of the smoothness of the deformation field without sacrificing the runtime advantage or registration accuracy.

\end{abstract}

\section{Introduction}
Deformable image registration and the subsequent quantitative assessment are crucial in a variety of medical imaging studies. Recent deep learning-based image registration (DLIR) methods \cite{yang2017quicksilver,krebs2017robust,de2017end,balakrishnan2018unsupervised,dalca2018unsupervised} have achieved remarkable results and showed immense potential for time-sensitive medical imaging studies such as image-guided surgery and motion tracking. Unsupervised DLIR methods \cite{balakrishnan2018unsupervised,hering2019mlvirnet,mok2020fast,mok2020large} circumvent costly iterative optimization in conventional image registration approaches by re-formulating the image registration problem as a learning problem with convolutional neural networks (CNN), resulting in fast image registration. While DLIR methods have a learning formulation that differs from the conventional image registration approaches \cite{thirion1998image,ashburner2007fast,avants2008symmetric,vercauteren2009diffeomorphic}, the tradeoff between registration accuracy and the smoothness of the deformation field, which is often controlled with a hyperparameter in the objective function, cannot be circumvented by DLIR methods. Typically, the optimal hyperparameter is determined using grid searching on the validation dataset \cite{balakrishnan2018unsupervised,mok2020fast}. Ironically, despite the runtime advantage of DLIR methods, searching for the optimal hyperparameter value is notoriously time-consuming and computationally intensive in DLIR methods as the hyperparameters are fixed throughout the learning and inference phase. In DLIR methods, each grid search value requires a new DLIR model trained with the distinct hyperparameter value, and each DLIR model requires up to $\sim 20$ hours to a few days to train from scratch \cite{balakrishnan2018unsupervised}. As such, analyzing the effect of hyperparameters and searching for optimal regularization parameters prove to be too prohibitive in DLIR methods, leading to suboptimal registration results and limited clinical applications. Despite the computational cost of the hyperparameter searching technique, the traditional hyperparameter searching technique may not be a good solution for unsupervised DLIR methods for two thoughtful reasons. First, the optimal regularization parameter is subject to the degree of misalignment between the input images, image modality, and intensity distribution. Second, the prior knowledge of the learned model cannot be utilized in the traditional hyperparameter searching technique, resulting in a substantial computational redundancy.

In recent years, a pioneering work of Gatys et al. \cite{gatys2016image} demonstrate that CNN encodes both the content and style information of an image. Subsequent studies \cite{dumoulin2016learned,huang2017arbitrary,chen2018on,karras2019style} further illustrate that the image information can be separated by manipulating the statistics of the feature maps with feature-wise linear modulation \cite{dumoulin2018feature} in CNN. In this paper, motivated by these studies \cite{dumoulin2016learned,huang2017arbitrary,karras2019style}, we propose a novel conditional image registration method and a new self-supervised learning paradigm for deformable image registration to address the inefficiency of existing hyperparameter searching technique in DLIR methods. Instead of training multiple models for searching the optimal hyperparameter, we propose utilizing a single conditional model with self-supervised learning for efficient hyperparameter tuning. 

Parallel to our work, Hoopes et al. \cite{hoopes2021hypermorph} propose to learn the effects of registration hyperparameters on deformation field with Hypernetworks \cite{ha2016hypernetworks}, which leverage a secondary network to generate the conditioned weights for the entire network layers. While the Hypernetworks-based method offers immense modulation potential, it adds an enormous number of parameters to the original image registration method. Alternatively, we propose a more parameter-efficient and scalable approach based on conditional instance normalization. Our method learns the effect of the regularization parameters and conditions on the feature statistics of high-dimensional layers such that the smoothness of the solution can be manipulated via arbitrary hyperparameter values during the inference phase. We further introduce a novel distributed mapping network to generate non-linear embedding with the condition variable. We present extensive experiments, demonstrating that our formulation enables the precise control of the smoothness of the deformation field during the inference phase and rapid grid search of an optimal hyperparameter without sacrificing the runtime advantage or the registration accuracy of the original DLIR method. 



\section{Methods}
Deformable image registration establishes a dense non-linear correspondence between a fixed image $F$ and a moving image $M$, and the solution $\phi$ is often subject to a weighted smoothness regularization. DLIR methods often formulate the deformable image registration problem as a learning problem $\phi = f_\theta(F, M)$, in which $f_\theta$ is parameterized with CNN. Therefore, in contrast to conventional image registration approaches, the strength of the smoothness regularization is fixed throughout the training and inference phase. To address this limitation, we extend the common formulation of DLIR methods to a conditional deformable image registration setting. Instead of learning to adapt a particular weighted smoothness regularization, our proposed method learns the conditional features that correlated with arbitrary hyperparameter values. In the following sections, we describe the methodology of our proposed method.

\begin{figure}[t]
	\centering
	\begin{tabular}{cc}
		\includegraphics[width=0.46\linewidth]{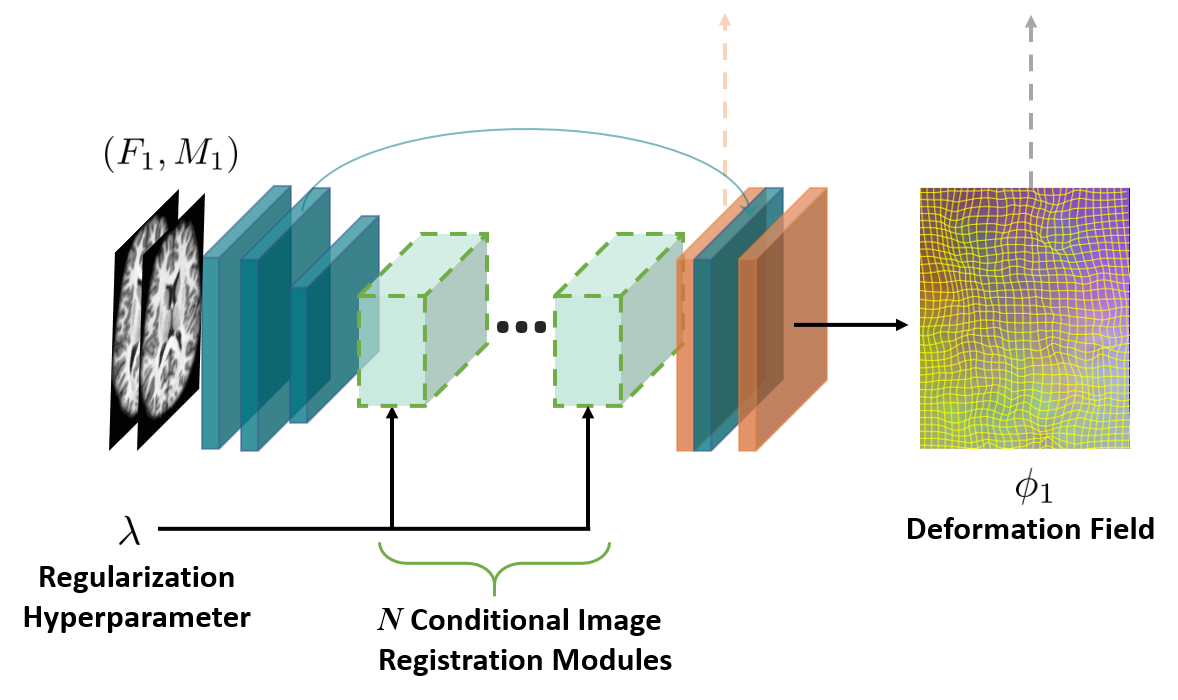} &
		\includegraphics[width=0.51\linewidth]{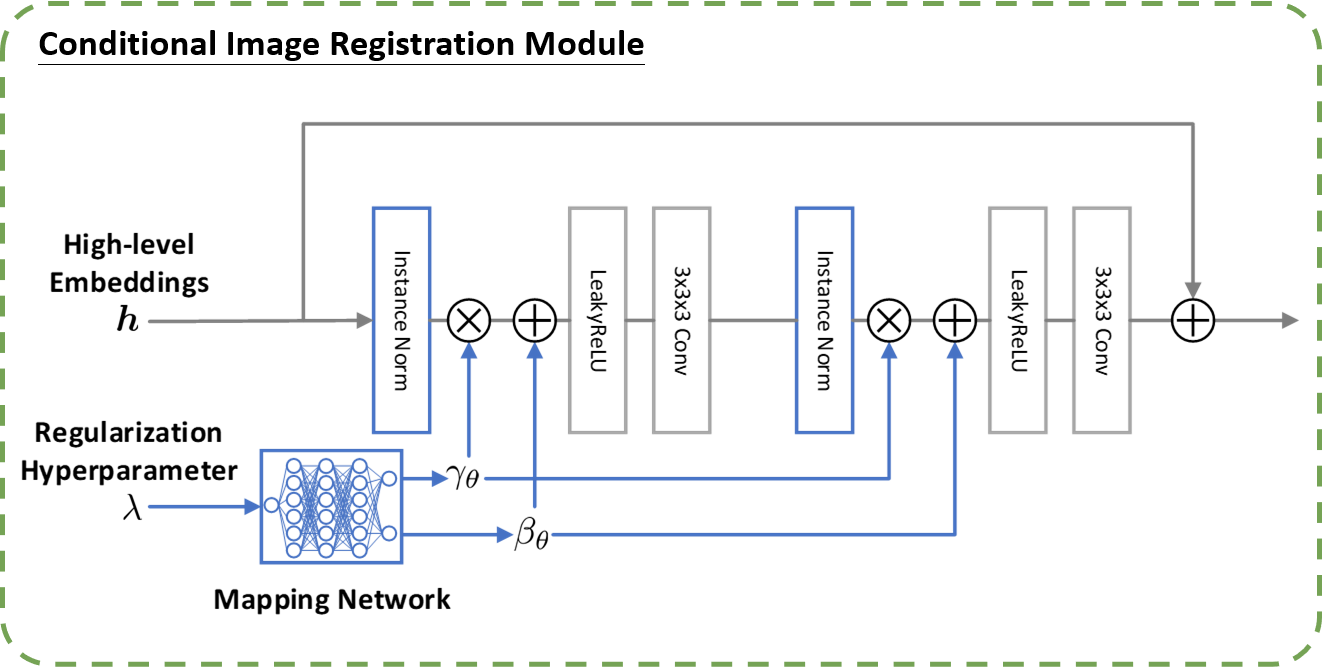} \\
		(a)   & (b)
	\end{tabular}
	
	\caption{Overview of the proposed (a) conditional deformable image registration method and (b) the conditional image registration module. For clarity and simplicity, we depict the first pyramid level only and illustrate the 2D formulation of our method in the figure.} \label{fig:FCN_archi}
\end{figure}

\subsection{Conditional Deformable Image Registration}

Given a fixed $F$, a moving 3D image scan $M$, and a conditional variable $c$, we parametrize the proposed condition image registration method as a function $f_\theta(F,M,c) = \phi$ with CNN. The proposed method works with any CNN-based DLIR methods and conditional variables. Specifically, we parametrize an example of the function $f_\theta$ with the deep Laplacian pyramid image registration network (LapIRN) and set the conditional variable to the smoothness regularization parameter $\lambda$. To condition a CNN model on a conditional variable, a concatenation-based conditioning approach \cite{mirza2014conditional,zhang2017stackgan,zhang2018stackganv2,dumoulin2018feature} in generative models is to directly concatenate the condition variable with the input image scans. However, based on our experiments, we observed that the concatenation-based conditioning approach cannot capture a wide range of regularization parameters and bias to a limited range of hyperparameter values.

Therefore, we depart from the concatenation-based conditioning approach and extend the feature-wise linear modulation approach \cite{chen2018on,karras2019style} instead. We condition the hidden layers on the regularization parameter directly. In particular, the network architecture of LapIRN is comprised of $L$ CNN-based registration networks (CRN). Each CRN consists of three major components: a feature encoder, a set of $N$ residual blocks, and a feature decoder. We replace the $N$ residual blocks with our proposed conditional image registration modules, as shown in figure \ref{fig:FCN_archi}(a). The feature encoder extracts the necessary low-level features for deformable image registration, while the feature decoder upsamples and outputs the targeted displacement fields. We only condition the hidden layers in each conditional image registration module on the hyperparameter of the smoothness regularization. We set $L$ and $N$ to 3 and 5 in our experiments, respectively.


\subsection{Conditional Image Registration Module}
Based on the assumption that the characteristics of the deformation field, i.e. smoothness, can be captured and separated by CNN, we design the conditional image registration module that takes input hidden feature maps and the regularization hyperparameter as input, and outputs hidden features with shifted feature statistics based on conditional instance normalization (CIN) \cite{dumoulin2016learned}. Specifically, the proposed conditional image registration module adopts the pre-activation structure \cite{he2016identity} and includes two CIN layers, each followed by a leaky rectified linear unit (LeakyReLU) activation \cite{maas2013rectifier} with a negative slope of 0.2 and a convolutional layer with 28 filters, as depicted in figure \ref{fig:FCN_archi}(b). A skip connection is added to preserve the identity of the features.

\subsubsection{Conditional Instance Normalization}
While the centralized mapping network \cite{karras2019style} generates a conditional representation with less memory consumption and computational cost, we argue that the effective representation of the hyperparameter should be diverse and adaptable to different layers in CNN. Chen et al. \cite{chen2018on} demonstrate that modulating layers with various depths of CNN results in inconsistent performance, which implies that hidden features of different depths hold distinct feature statistics and non-linearly correspondence to the latent code. 

To maintain diverse conditional representations of the hyperparameter for each hidden level, we propose to include distributed mapping networks that learn a separate intermediate non-linear latent variable for each conditional image registration module, which shared among all the CIN layers. Formally, given a normalized regularization hyperparameter $\lambda \in \bm{\bar{\lambda}}$, the distributed mapping network \mbox{$g: \bm{\bar{\lambda}}\to\mathcal{Z}$} first maps $\lambda$ to latent code $\bm{z} \in \mathcal{Z}$. Then, the CIN layers learn a set of parameters that specialize $\bm{z}$ to the regularization smoothness. The distributed mapping network is parameterized with a 4-layer multilayer perceptron (MLP). For simplicity, we set the number of perceptrons in each MLP layer and the dimensionality of the latent space to 64. The middle layers in the distributed mapping network use the LeakyReLU activation to further introduce the non-linearity into the latent code. The CIN operation for each feature map $\bm{h_i}$ is defined as 

\begin{equation}\label{eq:control_reg}
\bm{h'_i} = \gamma_{\theta,i}(\bm{z}) \left( \frac{\bm{h_i}-\mu(\bm{h_i})}{\sigma(\bm{h_i})} \right) + \beta_{\theta,i}(\bm{z}),
\end{equation}

\noindent where $\gamma_{\theta,i}, \beta_{\theta,i} \in \mathbb{R}$ are affine parameters learned from the latent code $\bm{z}$, and $\mu(\bm{h_i}), \sigma(\bm{h_i}) \in \mathbb{R}$ are the channel-wise mean and standard deviation of feature map $\bm{h_i}$ in channel $i$. In other words, the control of smoothness regularization is learned by normalizing and shifting the feature statistics of the feature map with corresponding affine parameters $\gamma_{\theta,i}$ and $\beta_{\theta,i}$ for each channel in the hidden feature map $\bm{h}$. 

\begin{figure}[t]
	\begin{center}
		\includegraphics[width=1.0\linewidth]{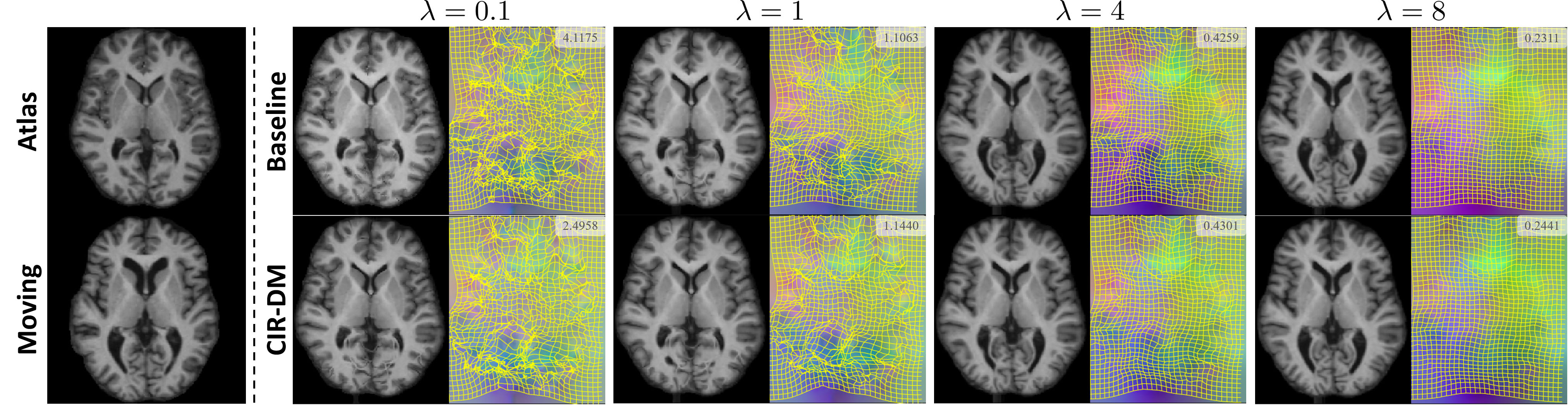}
	\end{center}
	\caption{Example axial MR slices of resulting warped images and deformation fields from the baseline method and our proposed method (CIR-DM) with $\lambda \in [0.1, 1, 4, 8]$. The standard deviation of the Jacobian determinant is shown at the upper-right corner of each resulting deformation fields.}
	\label{fig:qualitative}
\end{figure}

\subsection{Self-supervised Learning}
The objective of our proposed method is to compute the optimal deformation field corresponding to the hyperparameter of smoothness regularization. Formally, this task is defined as

\begin{equation}\label{eq:self_supervised}
    \phi^{*} = \argmin_{\phi} \mathcal{L}_{sim}(F, M(\phi)) + \lambda_p\mathcal{L}_{reg}(\phi),
\end{equation}
	
\noindent where $\phi^*$ denotes the optimal displacement field $\phi$, $\mathcal{L}_{sim}(\cdot,\cdot)$ denotes the dissimilarity function, $\mathcal{L}_{reg}(\cdot)$ represents the smoothness regularization function and $\lambda_p$ is uniformly sampled over a predefined range. We set the predefined range of $\lambda_p$ to $[0,10]$ empirically such that the optimal deformation field with maximum $\lambda_p$ is diffeomorphic in most cases. The only difference between the objective in common unsupervised DLIR methods \cite{balakrishnan2018unsupervised,hering2019mlvirnet,mok2020fast,mok2020large} and our objective is that we learn to optimize the objective function over a predefined range of hyperparameter instead of a fixed hyperparameter value. To exemplify our proposed learning paradigm, we follow \cite{mok2020large} and instantiate the objective function with a similarily pyramid and a diffusion regularizer on the spatial gradients of displacement fields. We also adopt a progressive training scheme to train the network in a coarse-to-fine manner. Mathematically, the objective function for each pyramid level $l \in L$ is defined as

\begin{equation}\label{eq:total_loss_example}
\mathcal{L}_l(F, M(\phi), \phi, \lambda_p) = \sum_{i \in [1 .. l]} -\frac{1}{2^{(l-i)}} NCC_w(F_i, M_i(\phi)) + \lambda_p||\nabla \phi||^2_2 ,
\end{equation}

\noindent where $\lambda_p$ is sampled uniformly in $[0,10]$ for each iteration and $NCC_w(\cdot,\cdot)$ denotes the local normalized cross-correlation (NCC) with window size $w$, in which $w$ is set to $1+2i$. It is worth noting that our proposed learning paradigm does not introduce extra computational cost to the original objective function and can be easily transferred to various DLIR applications with minimum efforts.


\section{Experiments}

\subsubsection{Data and Pre-processing}
We evaluate our method on brain atlas registration tasks. We use 425 T1-weighted brain MR scans from the OASIS \cite{marcus2007open,oasis_online} dataset and 40 brain MR scans from the LPBA40 \cite{shattuck2008construction,lpba_online} dataset. The OASIS dataset contains subjects aged from 18 to 96, and 100 of the included subjects were diagnosed with very mild to moderate Alzheimer's disease. We follow \cite{mok2020large} and perform standard pre-processing, including skull stripping, affine spatial normalization, intensity normalization, and subcortical structures segmentation, for each MR scan using FreeSurfer \cite{fischl2012freesurfer}. For the OASIS dataset, subcortical segmentation maps of 26 anatomical structures serve as the ground truth for the evaluation of our method. For the LPBA40 dataset, the brain MR scans in atlas space and its subcortical segmentation map of 56 anatomical structures, which are delineated by experts, are used in our experiments. We resample all MR scans with isotropic voxel sizes of $1^3 \text{mm}$ and center-cropped all the pre-processed image scans to $144 \times 192 \times 160$. We randomly split the OASIS dataset into 255, 20, and 150 volumes and split the LPBA40 dataset into 28, 2, and 10 volumes for training, validation, and test sets, respectively. We randomly select 3 and 2 MR scans from the test sets as atlases in OASIS and LPBA40, respectively. Finally, we register each subject to the chosen atlas using the baseline method and different conditional deformable image registration methods. In summary, there are 441 and 16 combinations of test scans from OASIS and LPBA40, respectively, included in the evaluation.

\begin{table}[t]
	\centering
	\caption{Quantitative results of the mean DSC and mean std($|J_\phi|$) over seven hyperparameter values on the OASIS and LPBA40 datasets. Initial: spatial normalization.}
	\label{tab:result}
	\resizebox{\textwidth}{!}{%
		\begin{tabular}{cccccccccccccc}
			\toprule[1.5pt]
			\multirow{2}{*}{Method} & \multicolumn{6}{c}{OASIS} & \multicolumn{6}{c}{LPBA40} \\
			\cmidrule(lr){2-7}\cmidrule(lr){8-13}
			& \rule{1pt}{0ex} DSC & \%DSC & std($|J_\phi|$) & \%std($|J_\phi|$) & $\textnormal{T}_{train}$ & $\textnormal{T}_{test}$ \rule{1pt}{0ex} & \rule{1pt}{0ex} DSC & \%DSC & std($|J_\phi|$) & \%std($|J_\phi|$) & $\textnormal{T}_{train}$ & $\textnormal{T}_{test}$ \rule{1pt}{0ex} \\
			\midrule[1pt]
			Initial \hspace{0.1cm} & 0.552 & - & - & - & - & - & 0.560 & - & - & - & - & - \\
			\midrule
			Baseline \hspace{0.1cm} & 0.770 & - & 1.157 & - & 200.3h & 0.204s & 0.729 & - & 0.697 & - & 143.2h & 0.206s  \\
			\midrule
			Traditional \hspace{0.1cm} & 0.780 & +1.41\% & 0.970 & +4.62\% & 28.4h & 0.212s & 0.722 & -1.01\% & 0.440 & -12.35\% & 20.3h & 0.210s \\
			\midrule
			CIR-CM \hspace{0.1cm} & 0.767 & -0.57\% & 0.900 & -5.23\% & 28.8h & 0.227s & 0.721  & -1.14\% & 0.473 & -5.73\% & 20.5h & 0.225s \\
			CIR-DM \hspace{0.1cm} & 0.770 & -0.19\% & 0.963 & -3.78\% & 28.5h & 0.216s & 0.728  & -0.17\% & 0.552 & -3.89\% & 20.4h & 0.218s\\

			\bottomrule[1.5pt]
		\end{tabular}
	}
\end{table}

\subsubsection{Implementation}
Our proposed method and the other baseline methods are implemented with PyTorch 1.7 \cite{paszke2017automatic} and deployed on the same machine, equipped with an Nvidia Titan RTX GPU and an Intel Core (i7-4790) CPU. We build our method on top of the official implementation of LapIRN available in \cite{offical_lapirn}. We adopt Adam optimizer \cite{kingma2014adam} with a fixed learning rate $0.0001$. We normalize $\bm{\bar{\lambda}}$ to [0,1]. We train all the methods from scratch (60000 iterations in OASIS and 40000 iterations in LPBA40). The source code will be published online.

\subsubsection{Baseline Methods}
We compare our method with the original LapIRN \cite{mok2020large} with a fixed hyperparameter (denoted as baseline). Specifically,  we train seven distinct LapIRNs with different regularization hyperparameters $\lambda \in [0.1, 0.5, 1, 2, 4, 8, 10]$. For each hyperparameter value $\lambda$, we select the top-3 models with the highest Dice score on the validation set for evaluation to alleviate the model variation. We further compare it with a concatenation-based conditioning approach (denoted as the traditional method) \cite{mirza2014conditional,zhang2018stackganv2,dumoulin2018feature}, which simply concatenates the regularization hyperparameter with the input scans in LapIRN to achieve conditional image registration. An ablation study of the variant of our proposed method is performed using either the 8-layer MLP centralized mapping network \cite{karras2019style} with latent space 256 (denoted as CIR-CM) and the proposed distributed mapping network (denoted as CIR-DM). For each condition deformable image registration method, we adopt the same training scheme and select the top-3 models with the highest Dice score ($\lambda = 0.1$) on the validation set for evaluation.

\subsubsection{Measurement}
We register each scan in the test set to an atlas, propagate the anatomical segmentation map of the moving image using the resulting deformation field with the nearest-neighbor interpolation, and measure the overlap of the segmentation maps using Dice similarity coefficient (DSC). We also measure the standard deviation of the Jacobian determinant on the deformation fields (std($|J_\phi|$)), representing the smoothness and local orientation consistency of the deformation field. Moreover, we compare each individual solution, which is generated with seven hyperparameter values $\lambda$, from all conditional methods to the solution of the corresponding test case generated from the baseline method, and measure the average difference (in percentage) of the mean Dice score (\%DSC) and the standard deviation of the Jacobian determinant on the deformation fields (\%std($|J_\phi|$)) over the total number of test cases. Finally, we measure the total training time in hours ($\textnormal{T}_{train}$) and the average inference time per case in seconds ($\textnormal{T}_{test}$) for each method. An ideal conditional image registration algorithm should achieve comparably registration accuracy and quality with the baseline method.


\begin{figure}[t]
	\centering
	\begin{tabular}{cc}
		\includegraphics[width=0.49\linewidth]{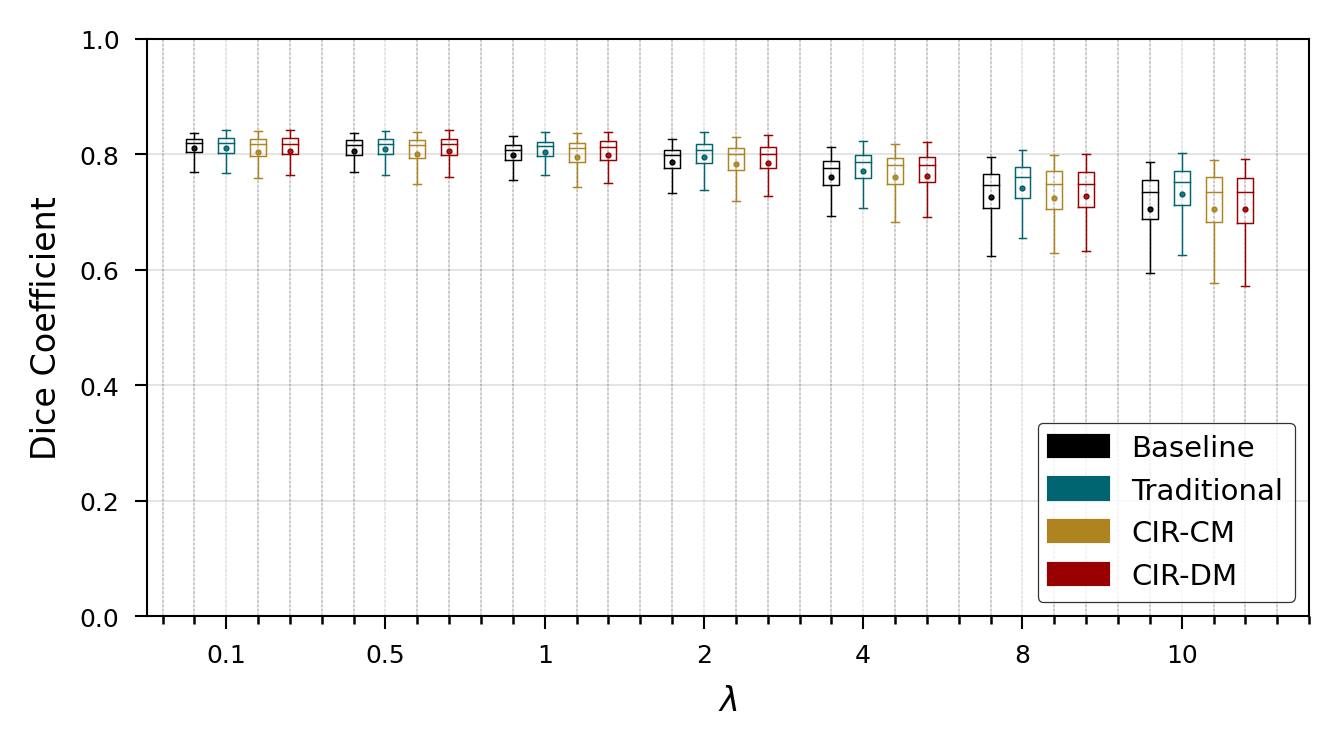} &
		\includegraphics[width=0.49\linewidth]{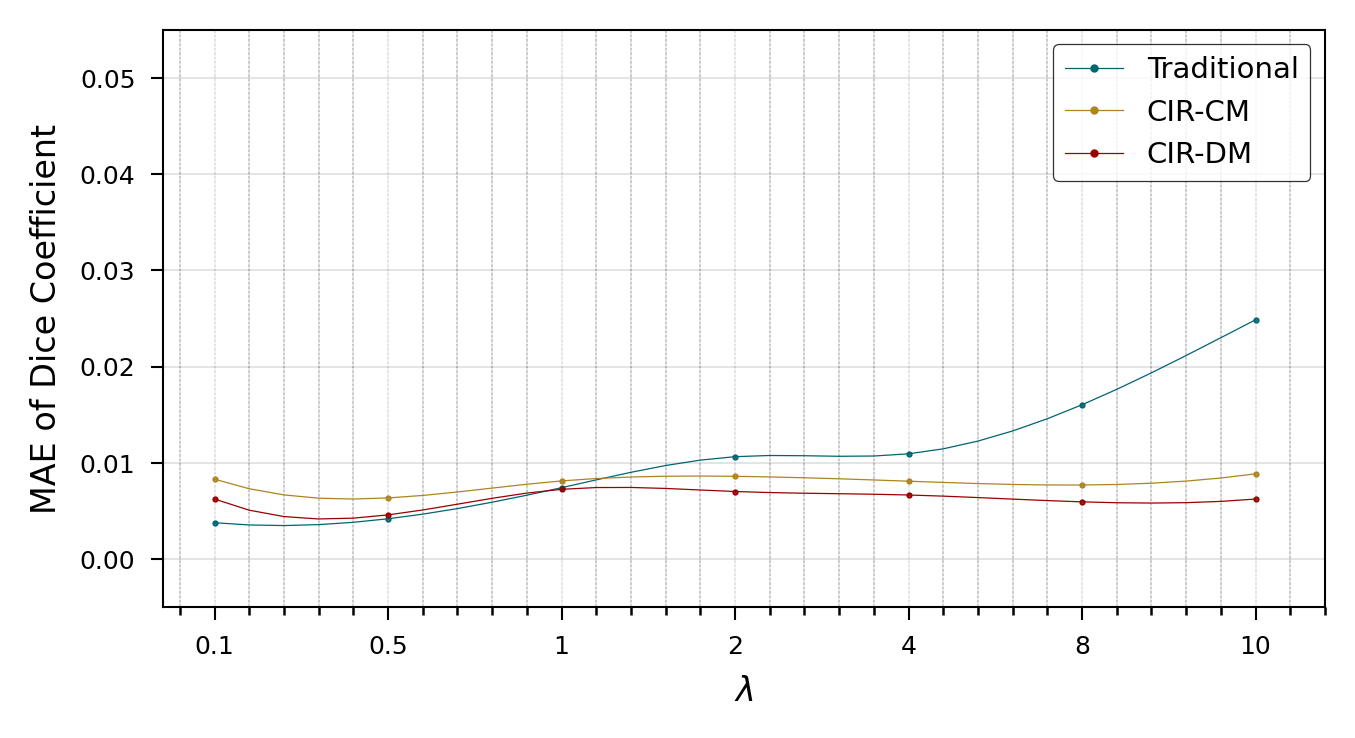} \\
		\includegraphics[width=0.49\linewidth]{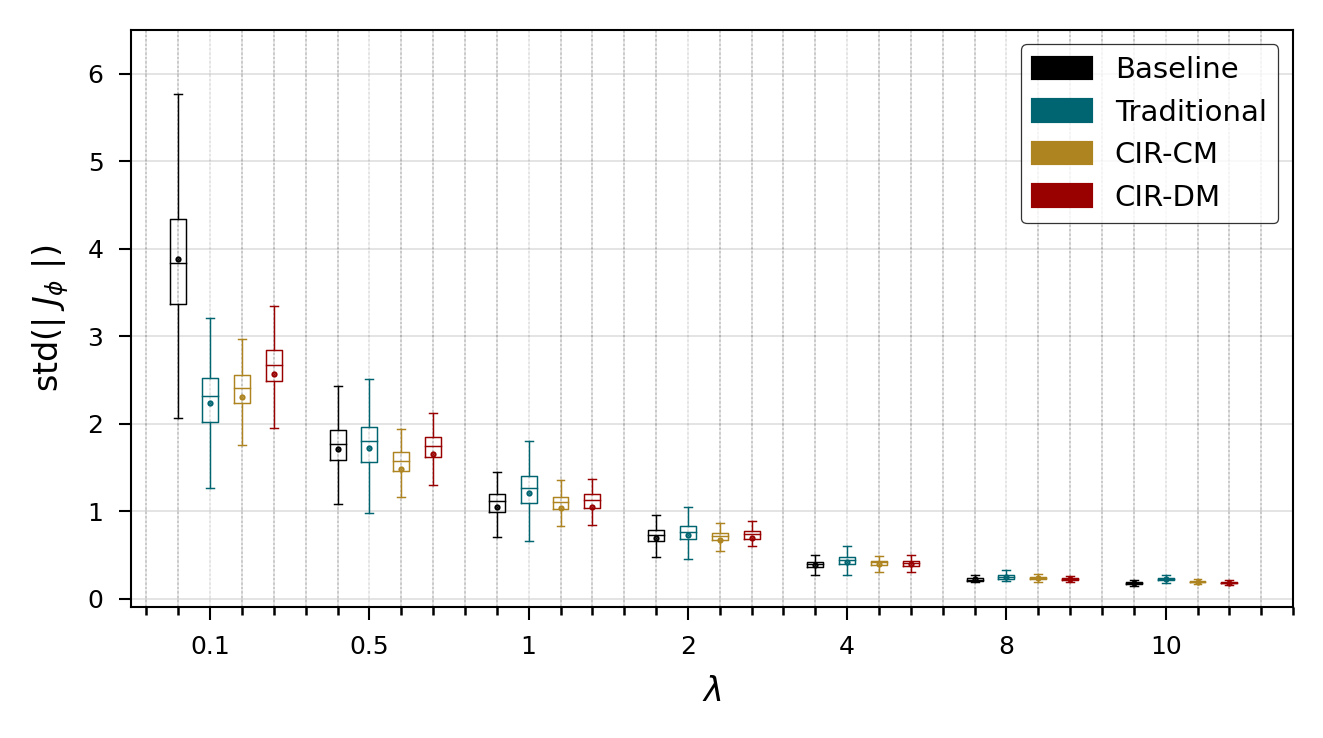} &
		\includegraphics[width=0.49\linewidth]{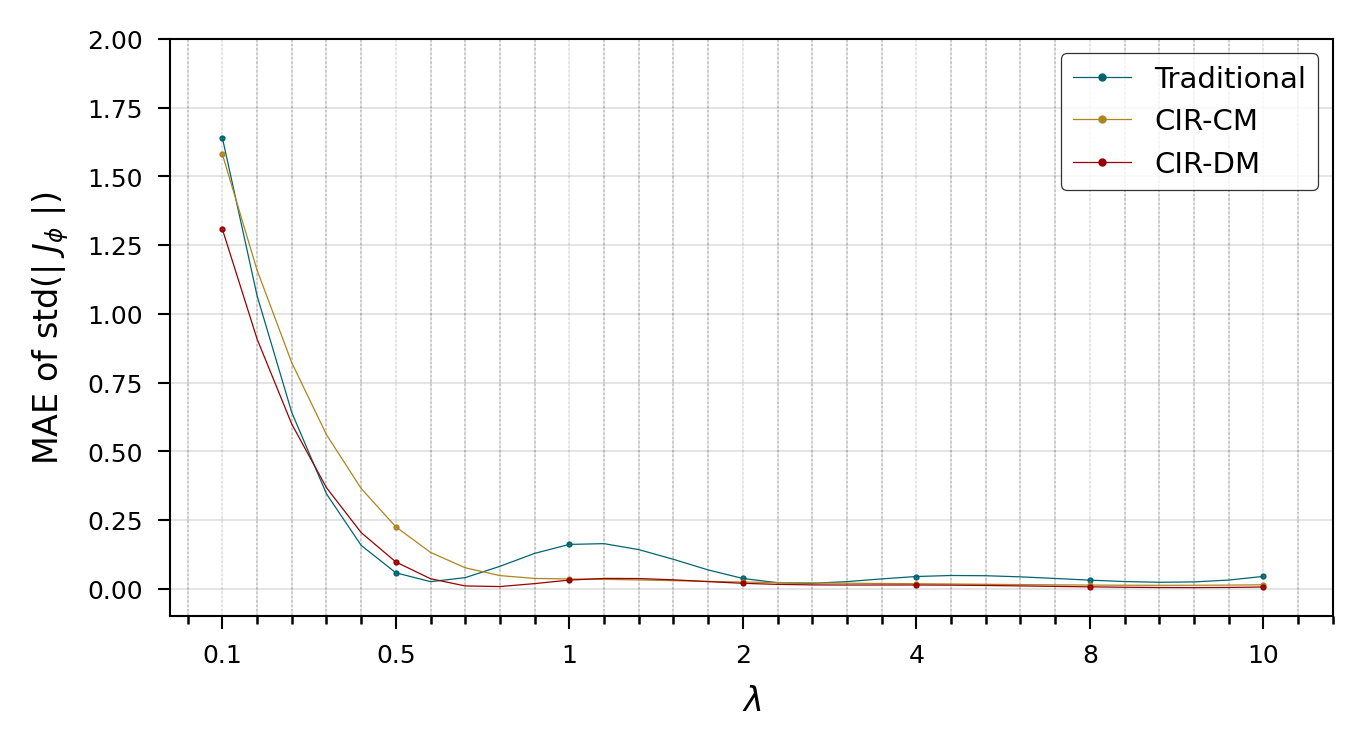} \\
	\end{tabular}
	\caption{Quantitative results over seven distinct hyperparameter values on the OASIS dataset. First row: the boxplot of Dice scores and the mean absolute error (MAE) of DSC compared to the baseline method. Second row: the boxplot of std($|J_\phi|$) and the MAE of std($|J_\phi|$) compared to the baseline method. The MAE of DSC (and the std($|J_\phi|$)) is computed by averaging the absolute difference of individual solutions between the targeting methods and baseline method over the total number of test cases.} \label{fig:quat_1}
\end{figure}

\subsubsection{Results and Discussions}
Table \ref{tab:result} presents a comprehensive summary of the results of each method in the OASIS and LPBA40 datasets. Figure \ref{fig:qualitative} illustrates qualitative results compare to the baseline method and figure \ref{fig:quat_1} shows detail results of each method over seven distinct hyperparameter values in the OASIS dataset. We demonstrate that not only does our method achieves highly consistent results with the baseline method, our method significantly reduces the total training time needed to generate solutions with diverse complexities. 

Specifically, all methods under our proposed conditional framework only required one trained model to generate solutions with seven distinct hyperparameter values of the smoothness regularization $\lambda$, resulting in $\sim$7x faster total training time than the baseline method. Interestingly, we find that the complexity of the resulting deformation fields (std($|J_\phi|$)) at $\lambda = 0.1$ declines significantly (-32\% to -41\%) while maintains comparable Dice scores with the baseline method, indicating that our methods produce even more desirable (smoother) solutions than the baseline method. In contrast to methods based on conditional instance normalization, the traditional method achieves a consistently higher average Dice score and standard deviation of the Jacobian determinant than the baseline method on the OASIS dataset when $\lambda \geq2$ as shown in figure \ref{fig:quat_1}, indicating the traditional method tends to bias to a limited range of $\lambda$. Compare to CIR-CM, our distributed mapping network design is in every way superior to the centralized mapping network in the context of conditional deformable image registration, as shown in figure \ref{fig:quat_1}. Importantly, our method achieves only -0.19\% (-0.17\% on LPBA40) difference of mean Dice score compared to the results of baseline method on OASIS, and the average inference time of CIR-DM is $\sim0.21$ seconds, highlighting the fact that CIR-DM is the only method that enables precise control of the deformation field regarding diverse $\lambda$ without sacrificing the registration accuracy or the runtime advantage of DLIR methods.

\section{Conclusion}
In summary, we have presented a novel conditional deformable image registration framework and self-supervised learning paradigm for deep learning-based deformable image registration. Our method learns the conditional features that are correlated with the regularization hyperparameter by shifting the feature statistics. It is demonstrated that our method enables precise control of the smoothness regularization in the inference phase without sacrificing the runtime advantage or the registration accuracy of the original DLIR method. Extensive experiments on brain atlas registration have been carried out, demonstrating that the results of our method consistently align with the results of the original DLIR method, and our method is superior to the common conditional approaches with diverse hyperparameter values. In principle, the proposed conditional image registration framework can be easily transferred to arbitrary CNN-based image registration approaches for controllable regularization of the deformation field and rapid hyperparameter tuning.

\bibliographystyle{splncs04}
\bibliography{myref}

\clearpage
\appendix
\begin{appendix}
	\renewcommand{\thesection}{\Alph{section}.}%
	
	\section{Quantitative results on the LPBA40 dataset}
	\begin{figure}[h]
		\centering
		\begin{tabular}{cc}
			\includegraphics[width=0.49\linewidth]{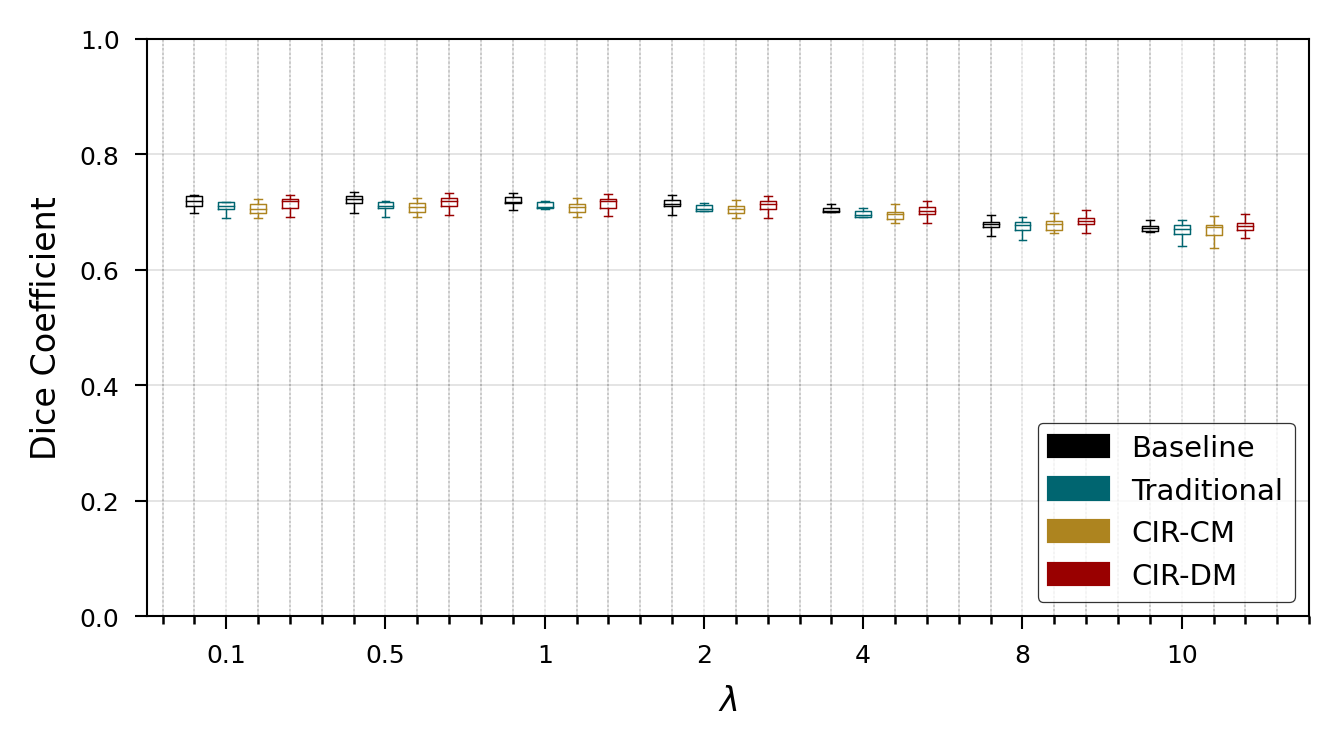} &
			\includegraphics[width=0.49\linewidth]{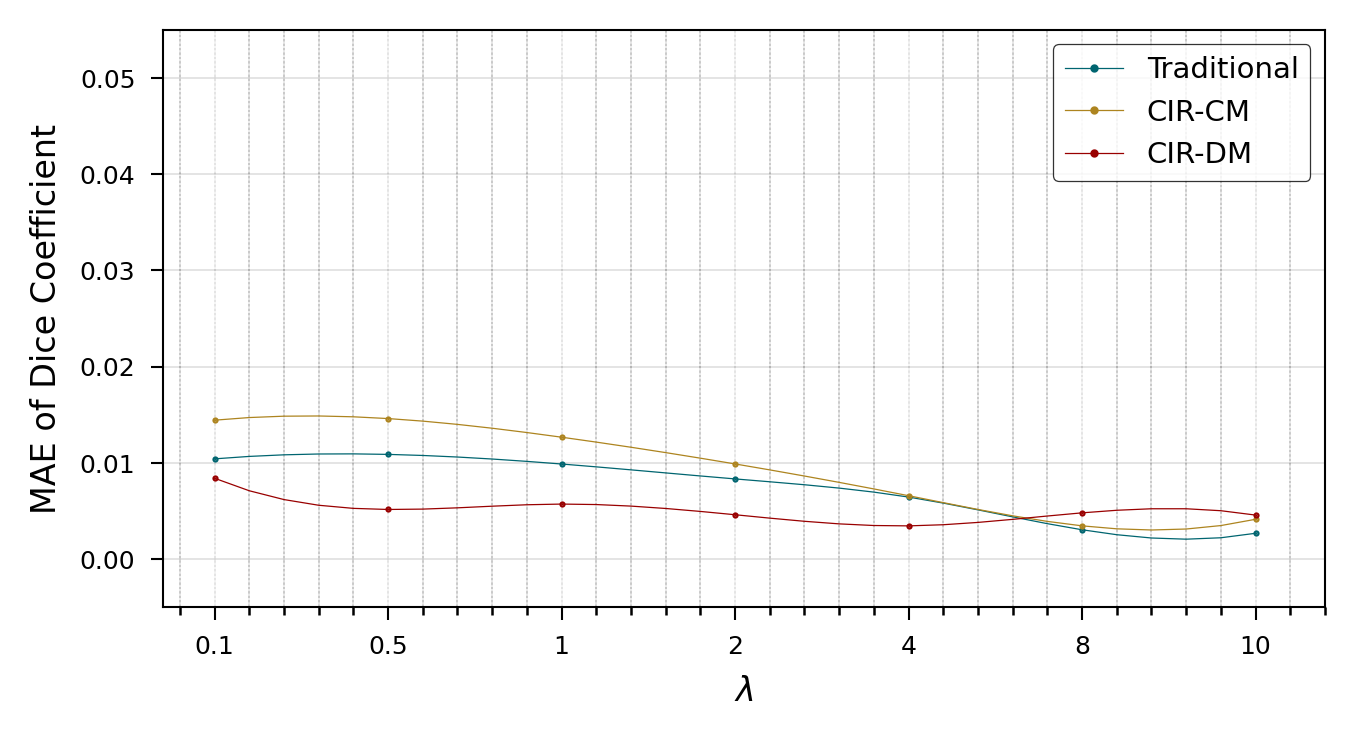} \\
			\includegraphics[width=0.49\linewidth]{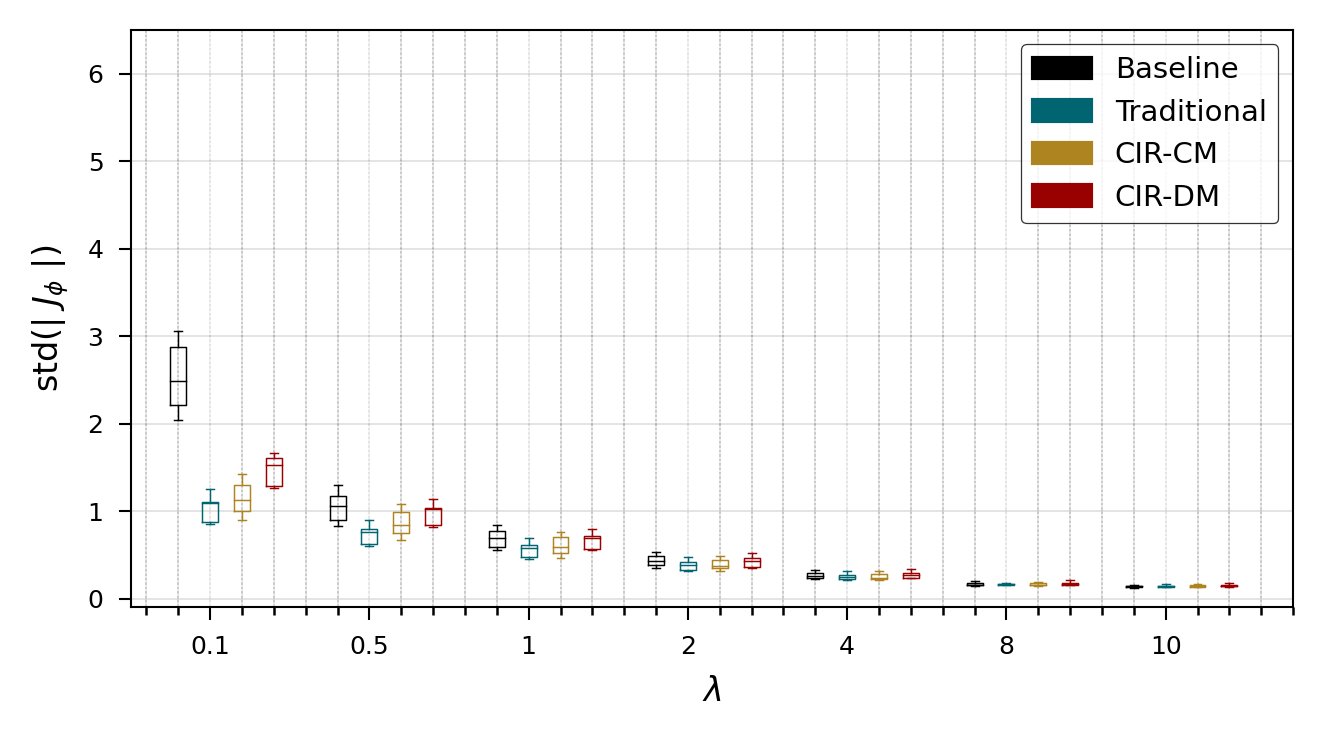} &
			\includegraphics[width=0.49\linewidth]{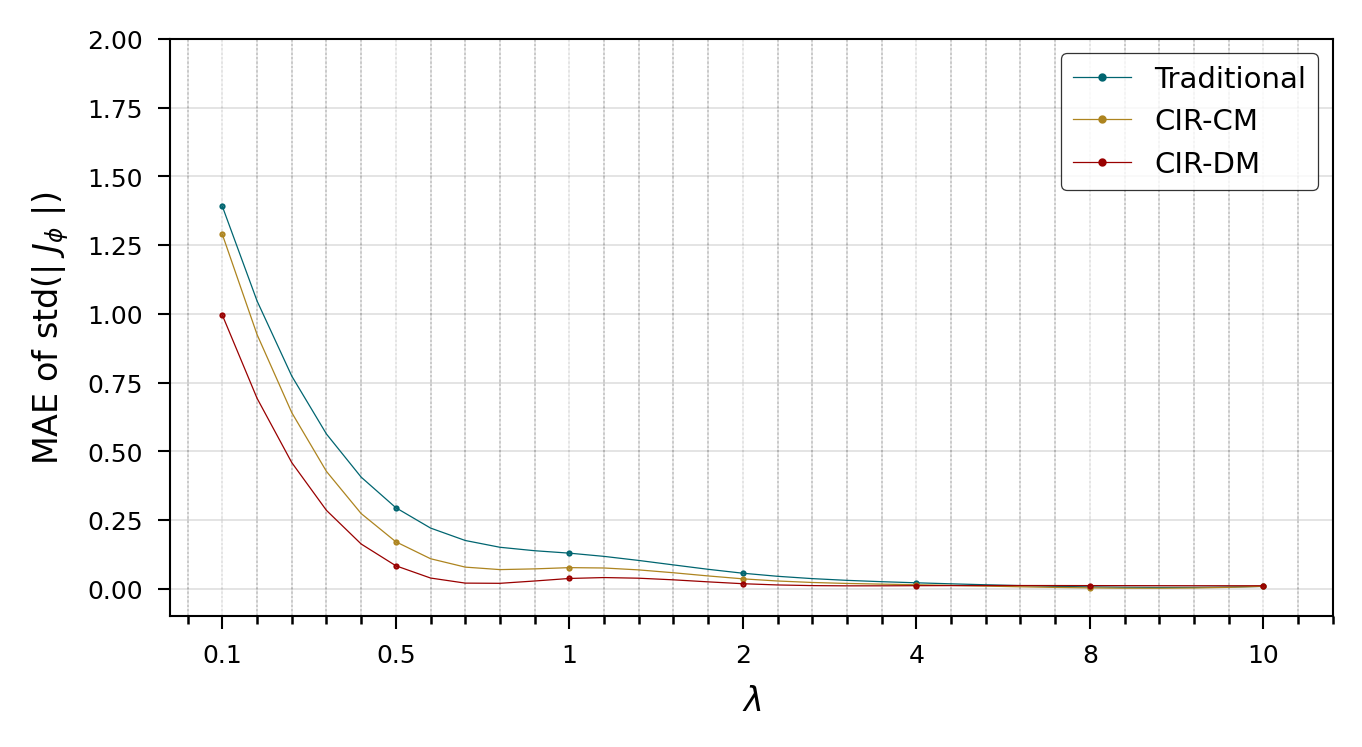} \\
		\end{tabular}
		\caption{Quantitative results over seven distinct hyperparameter values on the LPBA40 dataset. First row: the boxplot of Dice scores and the mean absolute error (MAE) of DSC compared to the baseline method. Second row: the boxplot of std($|J_\phi|$) and the MAE of std($|J_\phi|$) compared to the baseline method. The MAE of DSC (and the std($|J_\phi|$)) is computed by averaging the absolute difference of individual solutions between the targeting methods and baseline method over the total number of test cases.} 
		\label{fig:quat_1}
	\end{figure}
	
	\clearpage
	\section{Example MR slices in axial, sagittal and coronal plane}
	\begin{figure}[h]
		\begin{center}
			\includegraphics[width=1.0\linewidth]{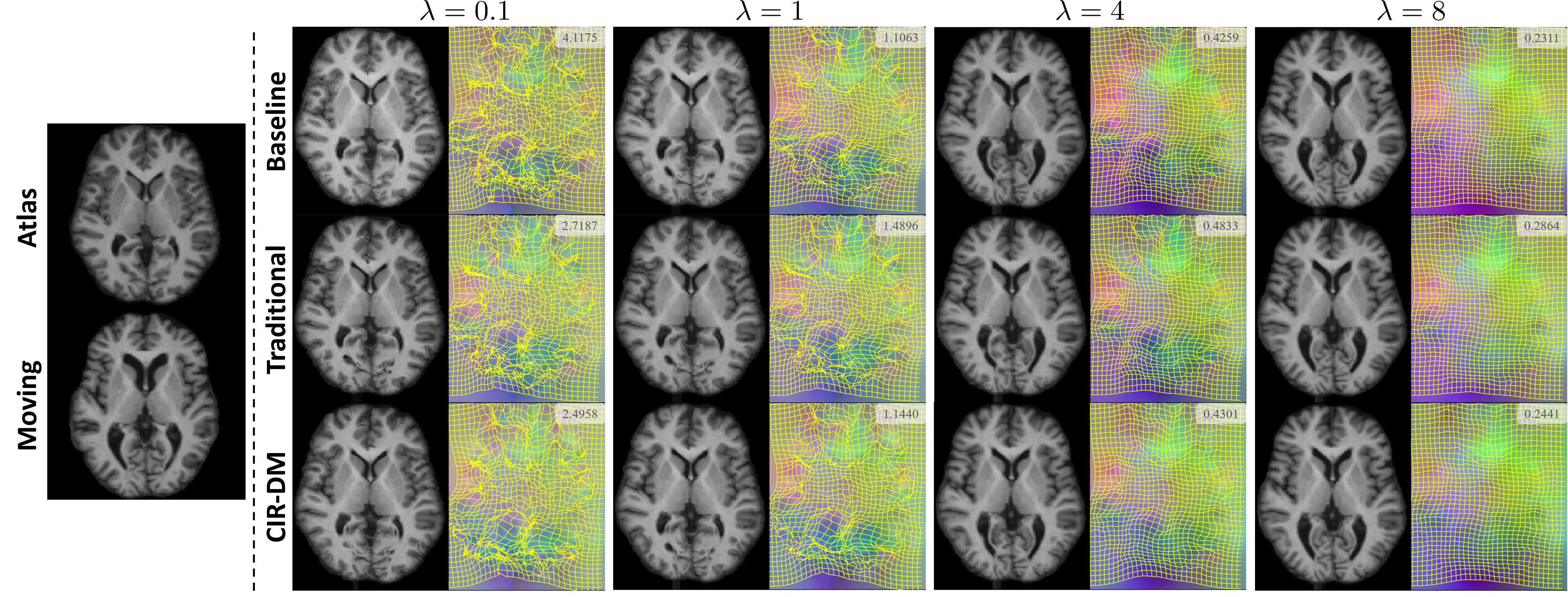}
		\end{center}
		\vspace{-12pt}
		\caption{Example axial MR slices of resulting warped images and deformation fields from the baseline method, traditional method and our proposed method (CIR-DM) with $\lambda \in [0.1, 1, 4, 8]$. The standard deviation of the Jacobian determinant is shown at the upper-right corner of each resulting deformation fields.}
		\label{fig:qualitative_a}
	\end{figure}
	
	\begin{figure}[h]
		\begin{center}
			\includegraphics[width=1.0\linewidth]{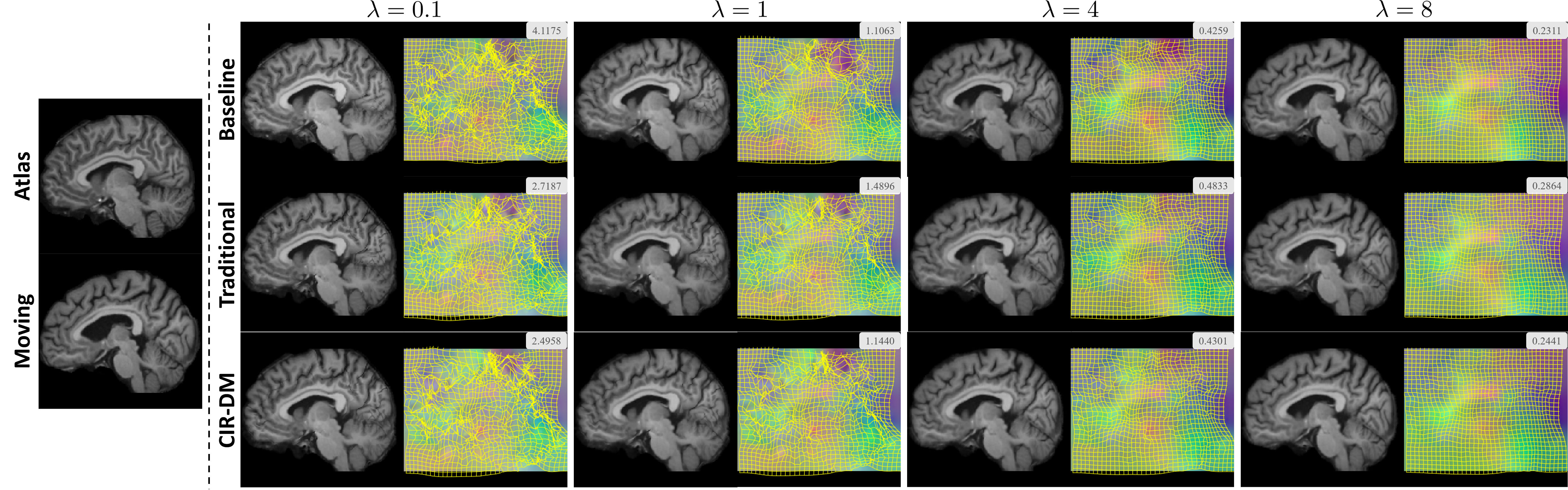}
		\end{center}
		\vspace{-12pt}
		\caption{Example sagittal MR slices of resulting warped images and deformation fields from baseline method, traditional method and our proposed method (CIR-DM) with $\lambda \in [0.1, 1, 4, 8]$. The standard deviation of the Jacobian determinant is shown at the upper-right corner of each resulting deformation fields.}
		\label{fig:qualitative_s}
	\end{figure}
	
	\begin{figure}[h]
		\begin{center}
			\includegraphics[width=1.0\linewidth]{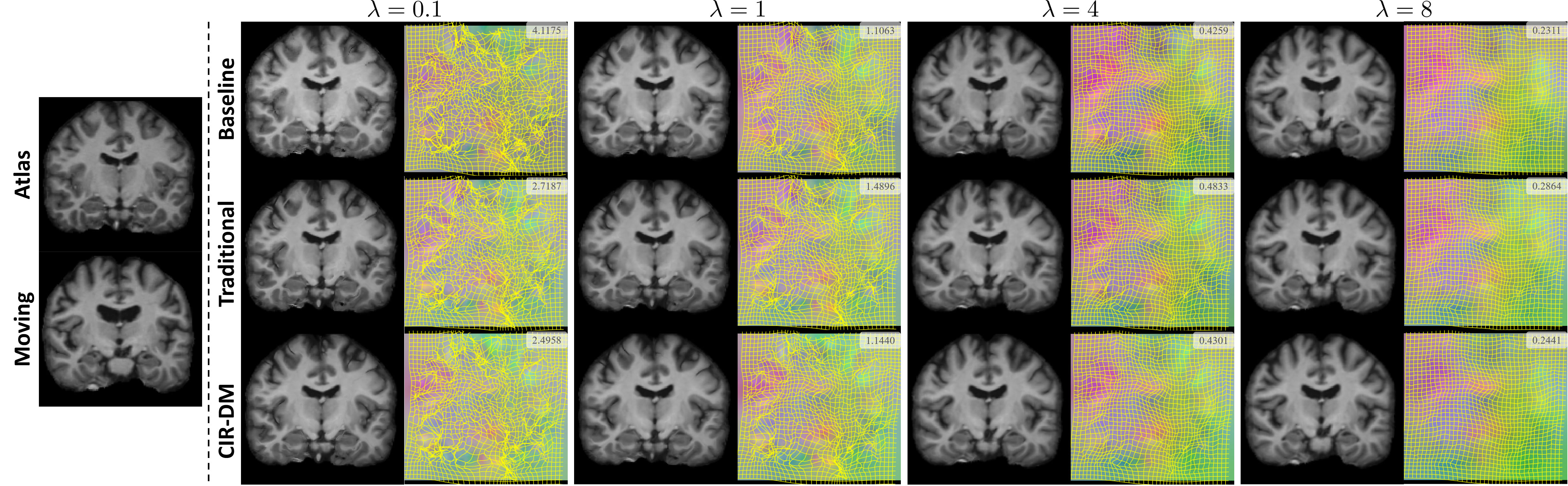}
		\end{center}
		\vspace{-12pt}
		\caption{Example coronal MR slices of resulting warped images and deformation fields from baseline method, traditional method and our proposed method (CIR-DM) with $\lambda \in [0.1, 1, 4, 8]$. The standard deviation of the Jacobian determinant is shown at the upper-right corner of each resulting deformation fields.}
		
		\label{fig:qualitative_c}
	\end{figure}
	
\end{appendix}

\end{document}